\documentclass[sn-mathphys-num,iicol]{sn-jnl}

\usepackage{graphicx}%
\usepackage{multirow}%
\usepackage{amsmath,amssymb,amsfonts}%
\usepackage{amsthm}%
\usepackage{mathrsfs}%
\usepackage[title]{appendix}%
\usepackage{xcolor}%
\usepackage{textcomp}%
\usepackage{manyfoot}%
\usepackage{booktabs}%
\usepackage{algorithm}%
\usepackage{algorithmicx}%
\usepackage{algpseudocode}%
\usepackage{listings}%
\usepackage[final]{changes}

\theoremstyle{thmstyleone}%
\theoremstyle{thmstyletwo}%

\theoremstyle{thmstylethree}%

\usepackage{soul,color,xcolor}      
\definecolor{myColor}{rgb}{0.8039,0,0}   
\makeatletter
\newcommand*{\revise}{\@ifnextchar\bgroup{\revise@}{\color{myColor}}}
\newcommand*{\revise@}[1]{{\textcolor{myColor}{#1}}}
\makeatother

\begin{document}

\title[Article Title]{High-Quality Spatial Reconstruction and Orthoimage Generation Using Efficient 2D Gaussian Splatting}


\author[1]{\fnm{Qian} \sur{Wang}}\email{qianwangcv@smail.nju.edu.cn}
\equalcont{These authors contributed equally to this work.}

\author[2]{\fnm{Zhihao} \sur{Zhan}}\email{zhzhan@topxgun.com}
\equalcont{These authors contributed equally to this work.}

\author[1]{\fnm{Jialei} \sur{He}}\email{jialeihe@smail.nju.edu.cn}

\author[1]{\fnm{Zhituo} \sur{Tu}}\email{xiaobeifenga@smail.nju.edu.cn}

\author*[1]{\fnm{Jie} \sur{Yuan}}\email{yuanjie@nju.edu.cn}

\affil[1]{\orgdiv{School of Electronic Science and Engineering}, \orgname{Nanjing University}, \orgaddress{\city{Nanjing}, \postcode{210046}, \country{China}}}

\affil[2]{\orgname{TopXGun Robotics}, \orgaddress{\city{Nanjing}, \postcode{211100}, \country{China}}}


\abstract{Highly accurate geometric precision and dense image features characterize True Digital Orthophoto Maps (TDOMs), which are in great demand for applications such as urban planning, infrastructure management, and environmental monitoring.
Traditional TDOM generation methods need sophisticated processes, such as Digital Surface Models (DSM) and occlusion detection, which are computationally expensive and prone to errors.
This work presents an alternative technique rooted in 2D Gaussian Splatting (2DGS), free of explicit DSM and occlusion detection. 
With depth map generation, spatial information for every pixel within the TDOM is retrieved and can reconstruct the scene with high precision. 
Divide-and-conquer strategy achieves excellent GS training and rendering with high-resolution TDOMs at a lower resource cost, which preserves higher quality of rendering on complex terrain and thin structure without a decrease in efficiency. 
Experimental results demonstrate the efficiency of large-scale scene reconstruction and high-precision terrain modeling.
This approach provides accurate spatial data, which assists users in better planning and decision-making based on maps.}




\maketitle

\section{Introduction}\label{sec1}
Digital Orthophoto Maps (DOMs) from Unmanned Aerial Vehicle (UAV) surveying provide dense textural and geometric data, useful in fields like agriculture, environmental monitoring, and disaster assessment~\cite{he2025multi}. 
DOMs are typically created by combining a Digital Elevation Model (DEM) with images captured from a fixed perspective, resulting in a nadir image of the target surface to correct projection distortion due to terrain and oblique photography.
Yet, occlusions of object facades can produce artifacts and incorrect geometry.
True Digital Orthophoto Maps (TDOMs) address this by incorporating Digital Surface Models (DSM) and using visibility checks to reduce occlusion effects and enhance map accuracy.

Over the past decades, numerous methods have been suggested for the generation of TDOMs. The Z-buffer method~\cite{article}, for example, resolves visibility by keeping track of distances between object points and perspective center corresponding to image pixels and selecting the nearest point. Recent learning-based approaches, such as GAN-based generation~\cite{shin2021true}, face limitations in generalization and often rely heavily on LiDAR data.

Neural Radiance Fields (NeRF)~\cite{mildenhall2021nerf} and 3D Gaussian Splatting (3DGS)~\cite{kerbl3Dgaussians}, emerging rendering-based methods, have introduced novel solutions for TDOMs. 
NeRF employs implicit 3D scene representation and differentiable rendering, but suffers from slow training and rendering speed.
Conversely, 3DGS explicitly encodes 3D scene geometry using Gaussian kernels and accelerates training and rendering through parallel processing, e.g., TOrtho-Gaussian~\cite{wang2024torthogaussiansplattingtruedigital}, achieving over 100 fps. 
However, 3DGS scalability is limited by GPU memory, and fidelity issues like blurring and aliasing may occur on reflections and thin structures.

This work describes a new TDOM-generation method based on a variant of 3DGS called 2DGS~\cite{Huang2DGS2024}. Surface reconstruction quality and perspective consistency are essential for TDOM generation. 2DGS addresses the perspective inconsistency of 3DGS by improving both geometric representation and rendering mechanisms. By leveraging normal information to refine surface structures, 2DGS yields higher-quality 3D reconstructions. Many TDOM applications require depth information to accurately represent 3D scene structure; our method extracts corresponding depth maps during rendering, thereby improving reconstruction accuracy. Inspired by VastGaussian~\cite{lin2024vastgaussian}, we adopt a divide-and-conquer strategy to enable large-scale 2DGS training and high-resolution TDOM rendering under limited computational resources.

\section{Related Works}\label{sec2}

\subsection{Traditional TDOM Generation Methods}
Traditional TDOM generation methods often require DSM priors and involve surface visibility analysis via occlusion detection and compensation using adjacent images.

The Z-buffer method~\cite{article} is one of the earliest visibility analysis algorithms. It is characterized by its simplicity but heavily depends on high-precision DSMs. Subsequently, the angle-based occlusion detection algorithm~\cite{habib2007new} was proposed, which detects occlusion by sequentially analyzing the angles of projected rays along radial directions in the DSM. Inspired by ray-tracing algorithms, Wang et al.~\cite{wang2009new} further optimized the algorithm to achieve smoother and more stable edge detection while reducing computational cost. Another approach is the vector polygon-based method~\cite{zhong2010vector}, which projects the vectors polygons representing building surfaces into the image space and evaluates occupancy priority in overlapping areas to acquire the actual coverage relationship.

Each of these methods has its own characteristics and specific application scenarios. However, they also come with inherent limitations. For instance, they rely heavily on high-quality DSMs or DBMs, which are often costly to acquire. Moreover, they still face challenges in accurate edge detection, natural texture compensation~\cite{shin2021true}, and achieving a streamlined and efficient workflow~\cite{zhou2020urban}. These limitations become even more pronounced in large-scale TDOM generation.

\subsection{Deep Learning-based Methods}
Recent methodological advancements have increasingly adopted deep learning frameworks to replace some intermediate steps in traditional TDOM generation workflows~\cite{shin2021true, ebrahimikia2024orthophoto}. For example, Urban-SnowflakeNet~\cite{ebrahimikia2024orthophoto} utilizes a deep learning network to extract features from preprocessed rooftop point clouds. It reconstructs building point clouds to aid in DSM rectification and TDOM generation, thereby effectively reducing distortions at building edges in urban scenes. However, these methods still rely on a combination of vision and LiDAR, requiring high-quality LiDAR point clouds and potentially failing in certain scenarios.

\begin{figure*}[ht]
    \includegraphics[width=1\textwidth]{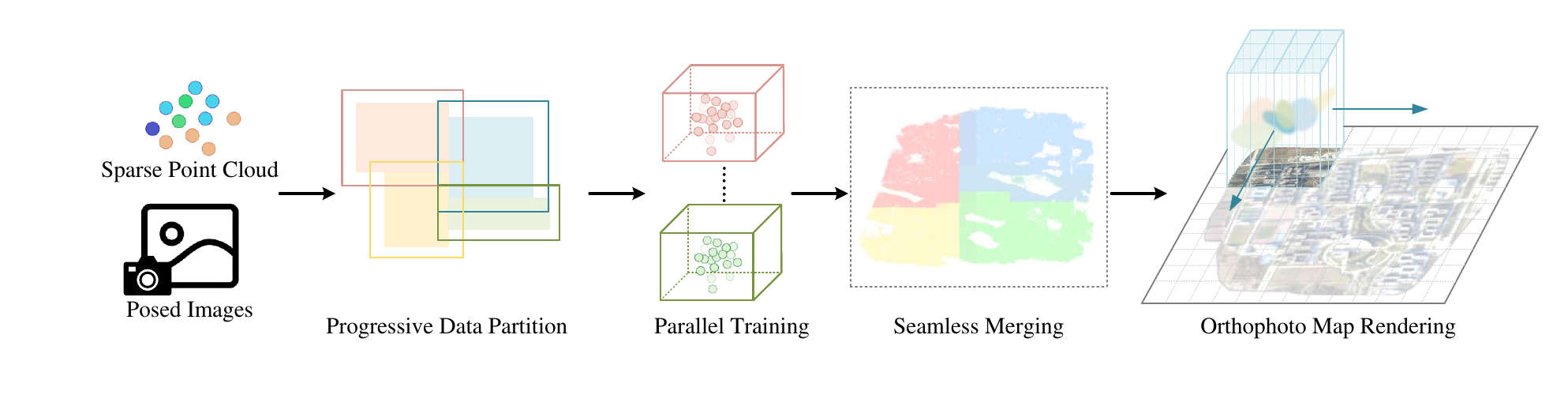}
    \caption{\textbf{Illustration of our pipeline.} 
    The input consists of sparse point clouds and images with poses. After progressive data partitioning, training is conducted in parallel on different GPUs. Eventually, the trained Gaussians are projected onto an image plane using an orthographic projection method, with the complete TDOM rendered through batch rasterization. Simultaneously, the corresponding depth images are generated.}
    \label{fig:pipeline}
    \vspace{-1.3em} 
\end{figure*}

\subsection{Differentiable Rendering-based Methods}
Novel view synthesis methods based on differentiable volume rendering have significantly improved the quality of reconstruction quality, providing a new paradigm for purely vision-based TDOM generation. Leveraging differentiable volume rendering, NeRF\cite{mildenhall2021nerf} learns an implicit neural radiance field, achieving remarkable fidelity and continuous scene reconstruction. Many speed-optimized NeRF variants have been applied to TDOM rendering, achieving image quality on par with traditional approaches~\cite{chen2024ortho}.

However, NeRF-based methods are limited by rendering efficiency. In contrast, 3DGS~\cite{kerbl20233d}, proposed by Bernhard et al., explicitly represents scenes with 3D Gaussian spheres and performs splat-based rendering via a high-speed differentiable rasterizer, enabling real-time, high-quality rendering. TOrtho-Gaussian~\cite{wang2024torthogaussiansplattingtruedigital}, a recent approach, addresses GPU memory limitations caused by the growing number of Gaussian spheres in large-scale scenes by employing a divide-and-conquer strategy. It avoids the complexity of traditional visibility analysis and shadow compensation while maintaining high computational efficiency, demonstrating great potential for 3DGS-based TDOM generation.

\section{Method}\label{sec3}
The overall TDOM production pipline is illustrated in Figure~\ref{fig:pipeline}. First, we initiate the point cloud as 2D Gaussians using the original 2DGS training strategy presented in Subsubsection~\ref{sec:preliminaries}. The orthographic camera model and the depth map generation method are introduced in Subsubsection~\ref{sec:orthocamera} and Subsubsection~\ref{sec:depthmap}. To support large-scale scenes, we adopt a divide-and-conquer strategy. Finally, the merged 2D Gaussians are rendered via batch rasterization to produce high-resolution TDOMs (Subsubsection~\ref{sec:trainingstrategy}).

\subsection{Preliminaries} 
\label{sec:preliminaries}
3DGS represents a scene using a set of 3D Gaussian ellipsoids $G(\mathbf{x})$, which are initialized from point clouds from Structure-from-Motion (SfM)\cite{schonberger2016structure}. Each 3D Gaussian primitive is parameterized by its mean $\mu$ and the covariance matrix $\mathbf{\Sigma}$:
\begin{equation}
G(\mathbf{x})=exp(-\frac{1}{2}(\mathbf{x}-\mu)^T\mathbf{\Sigma}(\mathbf{x}-\mu))
\end{equation}
where the covariance matrix $\mathbf{\Sigma}=\mathbf{RSS}^T\mathbf{R}^T$ is decomposed into a scaling matrix $\mathbf{S}$ and a rotation matrix $\mathbf{R}$. Moreover, the properties of Gaussian primitives also include the opacity $\alpha$ and the spherical harmonic (SH) coefficients. To render an image, 3D Gaussian primitives must be projected onto the 2D image plane. Given a viewpoint $V_k$, the projected 2D covariance matrix $\mathbf{\Sigma}_k$ in projection space is computed using the view transformation matrix $\mathbf{W}$ and the projection matrix $\mathbf{J}$: 
\begin{equation}
    \mathbf{\Sigma}_k=\mathbf{JW}\mathbf{\Sigma}\mathbf{W}^T\mathbf{J}^T
\end{equation}
By replacing the Gaussian center $\mu$ and the covariance matrix $\mathbf{\Sigma}$ with $\mu_k$ and $\mathbf{\Sigma}_k$, the corresponding 2D Gaussian primitive $G^{2D}$ on the image plane is obtained. After sorting the $N$ Gaussian primitives by depth, 3DGS renders the image from viewpoint $V_k$ using $\alpha$-blending:
\begin{equation}
    c(\mathbf{p})=\sum_{n=1}^N\alpha_nc_nG_n^{2D}(\mathbf{p})\prod_{i=1}^{n-1}(1-\alpha_iG_i^{2D}(\mathbf{p}))
\end{equation}
Here, $c_n$ represents the view-dependent appearance of the current Gaussian primitive $G_n^{2D}$ computed from spherical harmonics (SH) coefficients, and $c(\mathbf{p})$ denotes the color at pixel $\mathbf{p}$. The rendered results are used to compute the loss with respect to the ground truth and to derive gradients. Through backpropagation, the parameters of the Gaussian primitives are optimized accordingly.

2DGS flattens 3D Gaussian ellipsoids into 2D Gaussian disks. The planar shape of 2D Gaussians better aligns with surface reconstruction requirements and resolves depth inconsistency in 3DGS. This issue arises because the observed shape of Gaussian ellipsoids varies across different viewpoints, leading to inconsistent surface interpretation. 2DGS constructs a local coordinate system for each Gaussian. Any point $\mathbf{P}(u,v)$ in the local coordinate system can be transformed into world coordinates using the Gaussian center $\mu$, two principal tangential vectors $(\mathbf{t}_u, \mathbf{t}_v)$, and scaling factors $(s_u, s_v)$:
\begin{equation}
    \mathbf{P}(u,v)=\mu+s_u\mathbf{t}_uu+s_v\mathbf{t}_vv=\mathbf{H}(u,v,1,1)^T
\end{equation}
\begin{equation}
    where~\mathbf{H}=
    \begin{bmatrix}
        s_u\mathbf{t}_u&s_v\mathbf{t}_v&0&\mu\\
        0&0&0&1
    \end{bmatrix}
    =
    \begin{bmatrix}
        \mathbf{RS}&\mu\\
        \mathbf{0}&1
    \end{bmatrix}
\end{equation}
$\mathbf{R}=[\mathbf{t}_u,\mathbf{t}_v,\mathbf{t}_w]$ is a $3\times3$ rotation matrix and $\mathbf{S}=[s_u,s_v,0]$ is a $1\times3$ scaling matrix. Here, $\mathbf{t}_w = \mathbf{t}_u \times \mathbf{t}_v$ denotes the normal vector of the local plane. In the local coordinate system, the Gaussian primitive is defined as a standard 2D Gaussian.

To accurately compute ray–splat intersections, 2DGS proposes an intersection strategy that can be further used for depth estimation. Additionally, 2DGS introduces depth distortion loss and normal consistency loss to encourage Gaussian primitives to better conform to scene surfaces. Motivated by these advantages, this paper adopts the Gaussian kernel and loss functions proposed in 2DGS. For more details, we recommend readers refer to the original paper.

\subsection{Orthographic Projection of 2DGS}
\label{sec:orthocamera}
3DGS employs a pinhole camera model to perform perspective projection on the centers of Gaussian primitives, with all rays projected from the optical center of the camera, and 2DGS follows the same approach. Perspective projection defines a view frustum as the visible region and maps it into the normalized device coordinate (NDC) space within [-1,1]. The perspective projection matrix is given by:
\begin{equation}
    \mathbf{M}_{persp}=
    \begin{bmatrix}
        \frac{2z_n}{l-r}&0&\frac{r+l}{l-r}&0\\
        0&\frac{2z_n}{b-t}&\frac{t+b}{b-t}&0\\
        0&0&\frac{z_f+z_n}{z_f-z_n}&\frac{2z_fz_n}{z_f-z_n}\\
        0&0&1&0
    \end{bmatrix}
\end{equation}
\begin{equation}
    r=z_n\cdot tan(\frac{\theta_x}{2}),\quad
    t=z_n\cdot tan(\frac{\theta_y}{2})
\end{equation}
$z_n$ and $z_f$ represent the distances from the camera center to the near and far planes of the view frustum, respectively. $\theta_x$ and $\theta_y$ denote the horizontal and vertical field-of-view angles. The parameters $r, l, t,$ and $b$ correspond to the right, left, top, and bottom boundaries of the near plane.

\begin{figure}[ht]
    \centering
    \includegraphics[width=0.45\textwidth]{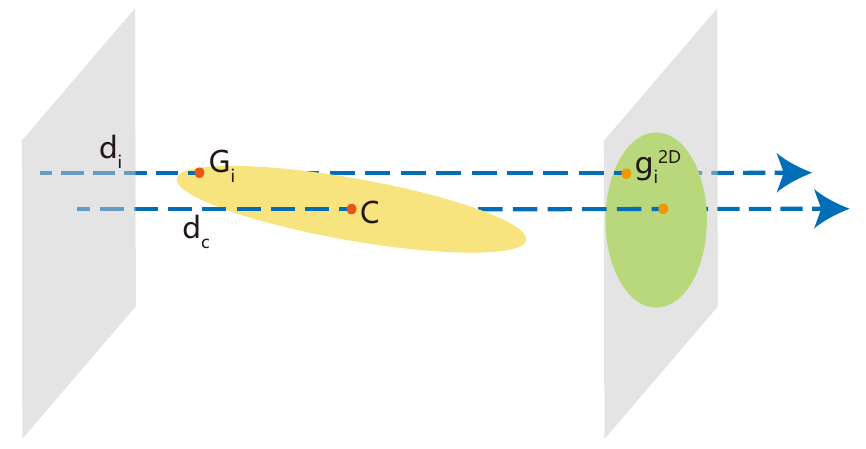}
    \caption{\textbf{Illustration of degenerate solutions in 2DGS.} 
    The Gaussian center is projected onto the image plane to construct a standard 2D Gaussian. Degeneracy in a given view is identified by comparing the Gaussian responses at the intersections between the current rays and the two Gaussian distributions.}
    \label{fig:degeneration}
    \vspace{-1.3em} 
\end{figure}

To obtain an orthographic image, we replace the perspective projection of Gaussian centers with orthographic projection. As shown in Figure~\ref{fig:degeneration}, in the orthographic camera model, all rays are parallel to the optical axis. Orthographic projection defines a cuboid view volume as the visible region, and the projection matrix is given by:
\begin{equation}
\mathbf{M}_{ortho}=
    \begin{bmatrix}
        \frac{2}{r-l}&0&0&\frac{r+l}{r-l}\\
        0&\frac{2}{t-b}&0&\frac{t+b}{b-t}\\
        0&0&\frac{2z_n}{z_n-z_f}&\frac{z_n+z_f}{z_n-z_f}\\
        0&0&0&1
    \end{bmatrix}
\end{equation}
Under a given viewpoint, each point is mapped from local space to image space via the transformation matrix $\mathbf{M}_{ortho}\mathbf{W}\mathbf{H}$. Unlike orthographic splatting in 3DGS, which requires an additional transformation of the covariance matrix, 2DGS eliminates the need for this extra step.

\subsection{Orthographic Depth Maps}
\label{sec:depthmap}
Generating an unbiased depth map corresponding to TDOM facilitates the extraction of terrain and semantic information from the scene. For each pixel, we calculate the expected termination depth of the corresponding ray using a formulation similar to volume rendering:
\begin{equation}
    D=\sum_{i=0}^{N-1}\alpha_iT_id_i
\end{equation}
\begin{equation}
    T_i=\prod_{j=0}^{i-1}(1-\alpha_{j})
\end{equation}
where $N$ denotes the count of Gaussians intersected by the ray, $\alpha_i$ represents the opacity at the ray-splat intersection $\mathbf{x}$, approximated by the Gaussian value $G_i(\mathbf{x})$ at the intersection, and $d_i$ denotes the distance from the origin of the ray to the intersection $\mathbf{x}$. In the orthographic model, all rays are parallel to the camera's optical axis. \deleted{The origin of each ray is designated as the intersection where the ray crosses a plane that passes through the camera optical center and is perpendicular to the ray.} Clearly, the value of $d_i$ is equivalent to the z-axis coordinate of the Ray-splat intersection point in the view space.

\replaced{Since 2D Gaussian primitives are elliptical disks, degeneracy may occur under certain viewing conditions. Specifically, when the viewing direction is nearly parallel to the plane of a Gaussian, its projected footprint on the image plane collapses, resulting in extremely small intersection responses. In this case, although a ray–splat intersection can still be computed, the Gaussian value $G_i(\mathbf{x})$ at the intersection becomes too small, making the resulting depth estimate unreliable. }{Since 2D Gaussian primitives are elliptical disks, when observed from a specific viewpoint, the primitives can degenerate from surfaces into lines, which is a degeneracy issue present in 2DGS. }Shown as Figure~\ref{fig:degeneration},\replaced{ we construct a standard 2D Gaussian $g_i^{2D}(\mathbf{x})$ on the image plane, centered at the projected Gaussian mean. If $g_i^{2D}(\mathbf{x}) > G_i(\mathbf{x})$,}{ When the Gaussian value $G_i(\mathbf{x})$ at the ray-splat intersection is too small, we create a standard 2D Gaussian distribution $g_i^{2D}(\mathbf{x})$, using the projection of the Gaussian primitive center on the image plane as the mean. By comparing the values at the intersections of the ray with the two Gaussians, when $g_i^{2D}(\mathbf{x})$ is greater than $G_i(\mathbf{x})$,} the 2D Gaussian for the current view is considered degenerate, and $d_i$ is taken as the z-axis coordinate of the Gaussian center in view space.

\subsection{High-Resolution TDOM Rendering Strategy}
\label{sec:trainingstrategy}
The scale and complexity of large scenes lead to an explosion in the number of Gaussian primitives used to represent the scene, coupled with the parallel rendering of each pixel, subsequently resulting in a high demand for GPU memory when rendering high-resolution DOMs of large scenes. \replaced{Inspired by Vast Gaussian, which also adopted in TOrtho-Gaussian, we employ a divide-and-conquer strategy by progressively partitioning the scene data during training. It is worth noting that this strategy is not a novel contribution of this work, but is incorporated as an effective mechanism for improving scalability within our pipeline.}{ Inspired by Vast Gaussian, we apply a divide-and-conquer approach, progressively partitioning the scene data during training.} The partitioning operation consists of three steps:

\begin{itemize}

\item \textbf{Camera position based scene partition}: Camera centers are projected onto the ground plane, and the scene is divided into $m \times n$ cells, ensuring an equal number of cameras per cell.
\item \textbf{Location based point selection}: The boundaries of each cell are expanded by a predefined ratio. The point cloud is projected onto the ground plane, and points within the expanded boundaries are selected as the input data for the current partition.
\item \textbf{Visibility based camera selection}: The visibility of the $j$-th cell to the $i$-th view is defined as $\Omega_{ij}$ / $\Omega_{i}$, where $\Omega_{i}$ denotes the image area of the $i$-th view and $\Omega_{ij}$ represents the projected area of the $j$-th cell's bounding box in the $i$-th view. If the visibility exceeds a threshold, the $i$-th camera will be chosen for the current cell.
Subsequently, the points covered by the i-th camera will be included in the cell.

\end{itemize}
The strategy of block-wise parallel training significantly reduces training time and mitigates the issue of insufficient GPU memory. To further address memory limitations, we adopt batch rasterization to generate orthographic panoramas with resolutions comparable to the input images. The output resolution is adjustable, depending on the size of the defined view volume.

\section{Experiment}\label{sec4}

In this section, we evaluate our TDOM generation method through both qualitative and quantitative comparisons with state-of-the-art commercial software and TOrtho-Gaussian. We also assess the quality of the generated depth maps.

\subsection{Experiments setup}

\begin{figure*}[ht]
    \centering
    \includegraphics[width=0.95\textwidth]{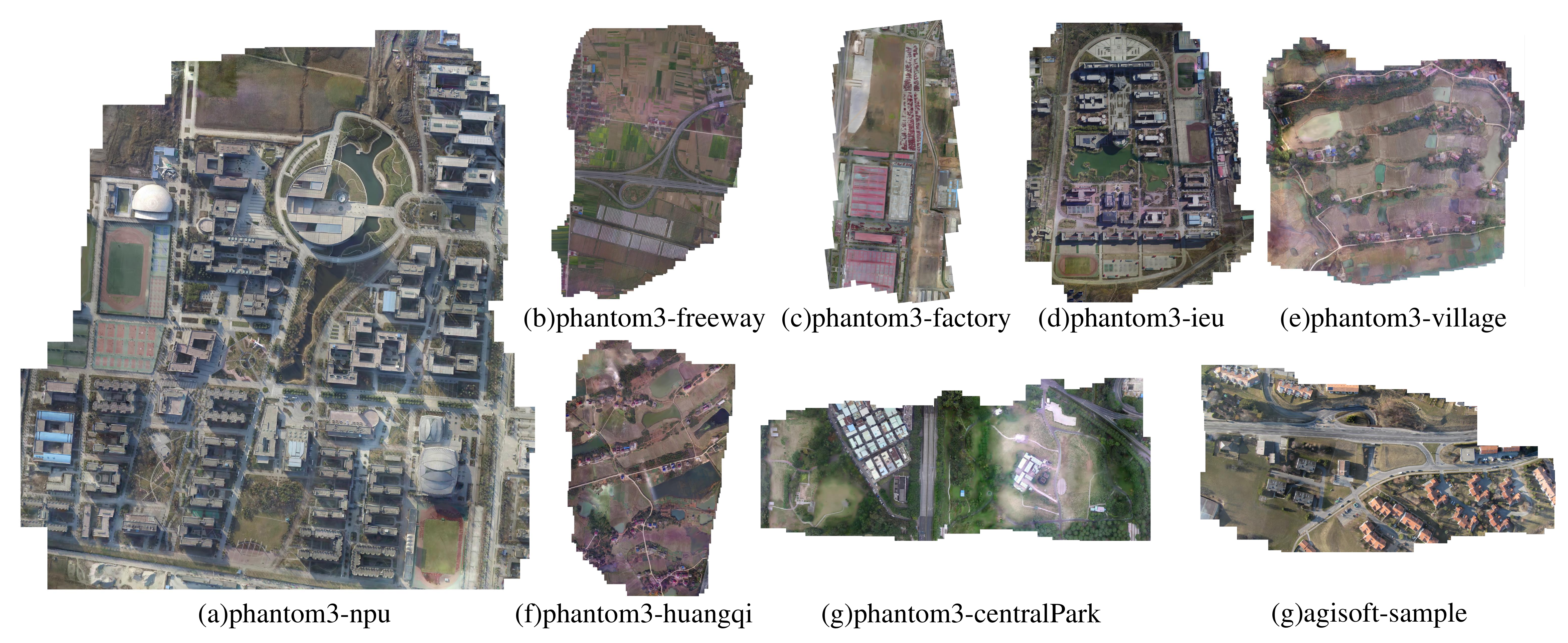}
    \caption{\textbf{Overview of TDOMs generated from the NPU DroneMap and Agisoft sample datasets using the proposed method.}}
    \label{fig:overview}
    \vspace{-1.3em} 
\end{figure*}

\textbf{Dataset:} We use several scenes with varying scales and characteristics from the NPU DroneMap Dataset\cite{bu2016map2dfusion}, where the largest scene covers an area of 1.598 km$^2$. These scenes cover diverse land use types, including urban high-rise buildings, rural areas, farmland, highways, and water bodies. The dataset was collected by Bu et al., using a custom-built hexacopter equipped with a GoPro Hero3+ camera and a DJI Phantom3. Additionally, we use a sample dataset provided by Agisoft~\cite{Agisortsampledata}, captured with a Canon DIGITAL IXUS 120 IS camera. The panoramic images generated by our method are shown in Figure~\ref{fig:overview}. \added{Table~\ref{tab:dataset} further summarizes the key statistics of these datasets, including the number of images, forward overlap, ground sample distance (GSD), spatial coverage, and image resolution.}

\textbf{Implementation:} The initial point cloud and camera poses obtained from COLMAP sparse reconstruction are first aligned using Manhattan alignment to match the scene bounding box with the coordinate axes. Each scene is partitioned into 4 subregions following the progressive partitioning strategy. Training is performed across four NVIDIA GeForce RTX 3090 GPUs for 60,000 iterations. After training, the Gaussians are filtered based on their central positions using the bounding boxes defined in camera-based partitioning. The average training time per scene is approximately 7 hours. Subsequently, the processed points are stitched and deduplicated for subsequent rendering. We adopt the orthographic camera model described in Subsection~\ref{sec:orthocamera} to render orthographic images. The full image is divided into $2 \times 2$ tiles and rendered sequentially. During rendering, the orthographic depth is computed simultaneously for each pixel.

\subsection{TDOM Evaluation}
We selected four commercial software, namely Map2D-Fusion~\cite{bu2016map2dfusion}, ContextCapture~\cite{contextcapture}, Metashape~\cite{metashape}, and Pix4Dmapper~\cite{pix4d}, which are based on traditional TDOM generation methods, as comparisons to evaluate our method. \added{All reconstruction methods, including COLMAP and commercial software, are executed using default configurations without manual tuning. In addition, we include TOrtho-Gaussian as a representative learning-based method for orthophoto generation to provide a more comprehensive comparison.} All the methods use the same input images and configurations. 

\begin{figure*}[ht]
    \centering
    \includegraphics[width=0.95\textwidth]{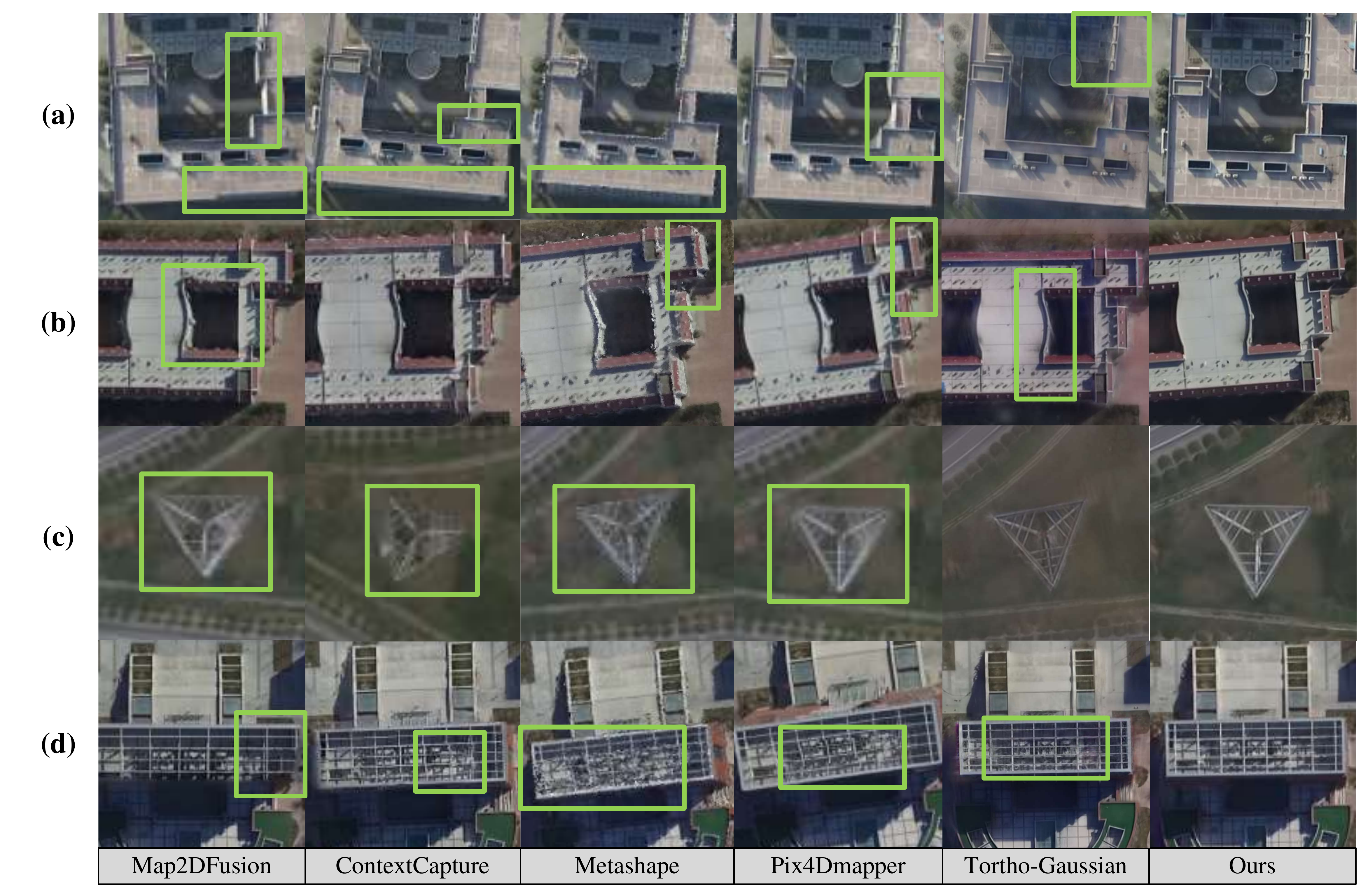}
    \caption{\textbf{Qualitative comparison with commercial software and learning-based methods on the NPU DroneMap dataset.} 
    Our method better preserves occlusion relationships and fine-grained scene details. Problematic regions are highlighted with green boxes for clarity.}
    \label{fig:comparison}
    \vspace{-1.3em} 
\end{figure*}

In the qualitative evaluation, we focused on buildings and thin structures.
\begin{itemize}
\item \textbf{Buildings:} For buildings, high-quality TDOM requires continuous, sharp edges and hidden facades. Our method excellently reconstructs the building boundaries and their occlusion relationships. In Figure~\ref{fig:comparison}(a) and Figure~\ref{fig:comparison}(b), it can be observed that other methods generally exhibit visible building facades. Map2DFusion shows the most severe issues due to direct photo mosaicking without image correction. Additionally, ContextCapture has blurred edges, Metashape displays noticeable irregular boundaries, and Pix4Dmapper suffers from misaligned building edges. \added{Tortho-Gaussian preserves overall structure but lacks high-frequency details, which degrade the visual quality and geometric consistency.}

\item \textbf{Thin structures:} Figure~\ref{fig:comparison}(c) and Figure~\ref{fig:comparison}(d) respectively demonstrate the restoration effects of the power tower and the air conditioner guardrail. Metashape consistently blurs scene information; the supports of the power tower become thickened and distorted in Map2DFusion, and are diminished in ContextCapture. The railings of the guardrail are significantly misaligned in all four software programs. In our method, these thin structures are clearer, and the details are well restored even under conditions of strong light and shadow contrast.
\end{itemize}

\begin{table*}[htbp]
\centering
\caption{\textbf{Overview of the datasets.}}
\small
\begin{tabular}{l c c c c c c}
\toprule
Sequence Name & Overlay & GSD (cm/pixel) & Area (km$^2$) & Image num. & Image size (pix) \\
\midrule
Phantom3-centralPark & 0.972 & 13.26 & 0.606 & 835 & 1920 $\times$ 1080 \\
Phantom3-factory & 0.901 & 16.29 & 0.912 & 402 & 1920 $\times$ 1080 \\
Phantom3-freeway & 0.924 & 21.17 & 1.457 & 415 & 1920 $\times$ 1080 \\
Phantom3-huangqi & 0.936 & 18.22 & 1.313 & 393 & 1920 $\times$ 1080 \\
Phantom3-ieu & 0.931 & 23.14 & 1.524 & 467 & 1920 $\times$ 1080 \\
Phantom3-npu$^*$ & 0.939 & 20.86 & 1.598 & 457 & 1920 $\times$ 1080 \\
Phantom3-village & 0.933 & 16.11 & 0.932 & 406 & 1920 $\times$ 1080 \\
\midrule
agisoft-sample & -- & -- & -- & 70 & 3000 $\times$ 4000 \\
\bottomrule
\end{tabular}
\label{tab:dataset}
\vspace{-1.3em} 
\end{table*}

\begin{table*}[htbp]
\centering
\caption{\textbf{Quantitative comparison of geometric accuracy using GCP-based RMSE (meters).}}
\small
\begin{tabular}{c|cccccc}
\toprule
 & Map2DFusion & Metashape & ContextCapture & Pix4Dmapper & TOrtho-Gaussian & Ours \\
\midrule
1 & 4.999 & \underline{3.258} & \textbf{1.228} & 5.460 & 3.455 & 3.572 \\
2 & 2.343 & 3.006 & \underline{1.166} & 2.146 & 1.247 & \textbf{1.013} \\
3 & 1.040 & \underline{1.442} & \textbf{0.785} & 2.272 & 1.740 & 1.724 \\
4 & 2.148 & \underline{1.293} & \textbf{0.382} & 2.308 & 1.371 & 1.430 \\
5 & 2.880 & 1.651 & \textbf{0.359} & 3.812 & 1.629 & \underline{1.545} \\
6 & 5.228 & \underline{2.642} & \textbf{0.288} & 6.795 & 2.576 & 2.285 \\
\midrule
RMSE & 3.106 & 2.215 & \textbf{0.701} & 3.798 & 2.003 & \underline{1.928} \\
\bottomrule
\end{tabular}
\label{tab:rmse}
\vspace{-1.3em} 
\end{table*}




\begin{table*}[htbp]
\centering
\caption{\textbf{Rendering metrics of TOrtho-Gaussian and our method.}}
\begin{tabular}{l ccc ccc}
\toprule
 & \multicolumn{3}{c}{Tortho-Gaussian} & \multicolumn{3}{c}{Ours} \\
\cmidrule(lr){2-4} \cmidrule(lr){5-7}
Scene & PSNR$\uparrow$ & SSIM$\uparrow$ & LPIPS$\downarrow$ 
      & PSNR$\uparrow$ & SSIM$\uparrow$ & LPIPS$\downarrow$ \\
\midrule
Phantom3-centralPark & \textbf{29.13} & \textbf{0.885} & \textbf{0.181} & 26.65 & 0.860 & 0.194 \\
Phantom3-factory & 29.05 & \textbf{0.877} & \textbf{0.218} & \textbf{29.29} & 0.873 & 0.251 \\
Phantom3-freeway & \textbf{32.64} & \textbf{0.908} & \textbf{0.164} & 30.26 & 0.864 & 0.231 \\
Phantom3-huangqi & \textbf{31.22} & \textbf{0.900} & \textbf{0.144} & 28.71 & 0.845 & 0.237 \\
Phantom3-ieu & \textbf{31.13} & 0.714 & \textbf{0.147} & 31.07 & \textbf{0.892} & 0.204 \\
Phantom3-npu & \textbf{30.94} & \textbf{0.926} & \textbf{0.133} & 29.32 & 0.910 & 0.180 \\
Phantom3-village & \textbf{32.01} & \textbf{0.906} & \textbf{0.175} & 29.00 & 0.828 & 0.288 \\
agisoft-sample & 19.54 & 0.501 & 0.514 & \textbf{25.91} & \textbf{0.847} & \textbf{0.188} \\
\midrule
Average & \textbf{29.46} & 0.827 & \textbf{0.210} & 28.78 & \textbf{0.865} & 0.222 \\
\bottomrule
\end{tabular}

\label{tab:img_quality}
\vspace{-1.3em} 
\end{table*}

In the quantitative evaluation, \deleted{the absolute distance error between two GCPs is applied as a criterion to evaluate the accuracy of the TDOM. We calculate the distances between the GCPs using both their true coordinate values and the coordinates extracted from the local coordinate system of the TDOM, while the true horizontal distances between the GCPs are computed using the Haversine formula. By aligning one pair of GCPs, we obtain a scaling factor, and then calculate the error between the aligned GCP distances from the TDOM and their true values. We compare our results with those from three commercial software solutions: Metashape, ContextCapture, and Pix4Dmapper. All the methods aligned the pair of control points GCP-101 and GCP-102. In our method, the minimum absolute error for GCP pairs is 0.267m, and the maximum absolute error is 2.736m. Our method maintains errors within 1.0m in the majority of cases and demonstrates comparable or superior accuracy relative to commercial software across most point pairs. Specific data can be found in Table~\ref{tab:accuracy}.} \added{since different software may use inconsistent internal coordinate systems, a unified evaluation is conducted by aligning manually labeled image coordinates with Ground Control Points (GCPs) from the Phantom3-npu dataset via a 2D similarity transformation in the ENU coordinate system. Specifically, all GCPs are first converted from geodetic coordinates into a local ENU coordinate system. A best-fit plane is estimated via singular value decomposition (SVD) and aligned to the horizontal plane. Manually labeled pixel coordinates are then aligned to the corresponding ENU coordinates using a 2D similarity transformation. After alignment, the Root Mean Square Error (RMSE) is computed in the ENU plane between the transformed pixel coordinates and the ground truth GCP positions.}

\added{
Table~\ref{tab:rmse} reports the per-point errors and overall RMSE. As shown, ContextCapture achieves the highest accuracy with an RMSE of 0.701 m. Our method achieves an RMSE of 1.928 m, slightly outperforming TOrtho-Gaussian (2.003 m), and is better than Metashape (2.315 m), Map2DFusion (3.106 m) and Pix4Dmapper (3.798 m). These results demonstrate that our method achieves competitive geometric accuracy, while maintaining performance comparable to traditional photogrammetric pipelines.
}

\added{In addition to geospatial accuracy, we assess image reconstruction quality via PSNR, SSIM, and LPIPS for both TOrtho-Gaussian and our method, with results shown in Table~\ref{tab:img_quality}. Our method achieves comparable SSIM to TOrtho-Gaussian, while PSNR and LPIPS are slightly lower. Under the single-view-dominant aerial setting and without additional constraints, rendering quality on unseen views is expected to degrade. Our main focus, however, is on 2DGS's capacity to represent planar surfaces, which is crucial for enhancing orthophoto accuracy.
}

\begin{figure}[ht]
    \centering
    \includegraphics[width=0.48\textwidth]{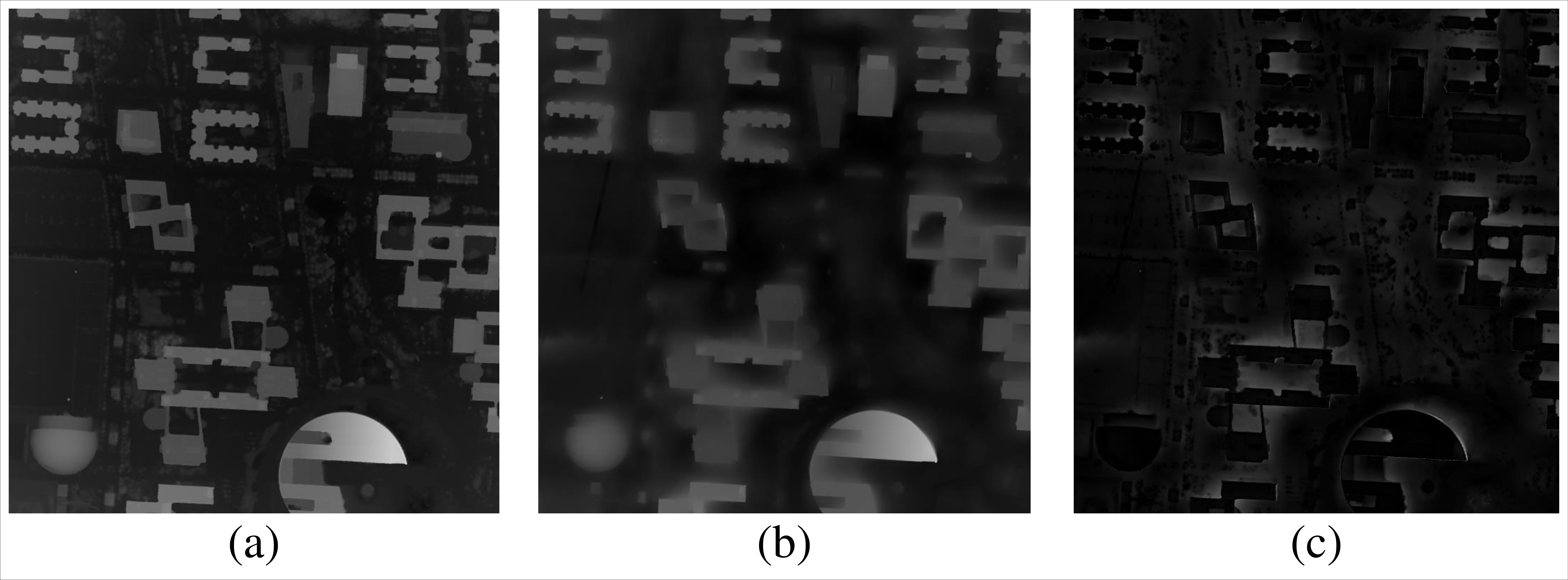}
    \caption{\textbf{Depth error visualization on Phantom3-npu dataset.} 
    (a) DEM grayscale map generated by Metashape; (b) DEM grayscale map produced by our method; (c) normalized depth error map.}
    \label{fig:depth_eval}
    \vspace{-1.3em} 
\end{figure}

\begin{table*}[ht]
\centering
\caption{\textbf{Quantitative DEM evaluation across four scenes using MAE (meters).} Metashape DEM is used as ground truth.}
\begin{tabular}{c|cccc}
\toprule
Scene & Phantom3-factory & Phantom3-huangqi & Phantom3-npu & Phantom3-village \\
\midrule
MAE (m) & 2.413 & 4.224 & 3.813 & 4.810  \\
\bottomrule
\end{tabular}

\label{tab:dem_mae}
\vspace{-1.3em}
\end{table*}


\subsection{Depth Map Evaluation}
\deleted{We synthesize the depth map by calculating the termination depth of the ray corresponding to each pixel in the orthophoto using a method similar to volume rendering. In this section, we will qualitatively evaluate the consistency between our depth map and the TDOM. }

\deleted{In the qualitative experiments, we replace the Red channel of the orthophoto with the depth map to facilitate an intuitive visualization of the correspondence between the depth information and the image. Additionally, we use the Canny edge detection algorithm~\cite{canny1986computational} to extract building edges from the depth map and overlay them onto our generated TDOM to observe the alignment of the edges. }

\deleted{As illustrated in Figure~\ref{fig:depth_eval}(1), The deeper red hue in the pixel indicates the higher height of the relevant surface object relative to the ground. the deeper red hue indicates higher elevation relative to the ground. The red areas predominantly cover the buildings on the surface, which aligns with real-world scene geometry. }

\deleted{In Figure~\ref{fig:depth_eval}(2), the edges extracted from the depth map distinctly outline the boundaries of the structures within the TDOM, as well as some of the taller trees on the terrain. The good alignment between depth edges and image structures indicates that the reconstructed depth preserves sharp geometric discontinuities and is consistent with the underlying scene layout. }

\deleted{Furthermore, compared to other methods, our depth map shows improved visual consistency with fewer noticeable artifacts and better preservation of high-frequency structural details, particularly along building boundaries and thin structures. }

\deleted{Our depth map adds a third dimension of height information to the TDOM, which is beneficial for subsequent tasks such as image detection and semantic segmentation in TDOM. }

\added{In this section, we evaluate the accuracy of the reconstructed depth maps by comparing them against the digital elevation model (DEM) generated by \textit{Metashape}, which is used as the ground truth. To ensure a consistent world coordinate system, we directly adopt the optimized camera poses and sparse point cloud provided by Metashape as the reference.}

\added{We quantitatively measure the accuracy using the mean absolute error (MAE) between our estimated DEM and the Metashape-derived DEM across multiple scenes. The results are summarized in Table~\ref{tab:dem_mae}.}

\added{As shown in Table~\ref{tab:dem_mae}, our method achieves consistently low DEM errors across all evaluated scenes. We also provide a qualitative visualization of the error distribution. As shown in Figure~\ref{fig:depth_eval}, the majority of error is concentrated along object edges, while planar regions remain highly consistent with the ground truth DEM. This suggests that our method preserves overall scene geometry well, while minor inaccuracies remain in high-frequency structural details.}

\added{Overall, the results demonstrate that our reconstructed depth maps are geometrically consistent with the Metashape DEM, both quantitatively and visually.}

\section{Conclusion}\label{sec5}
In this study, we propose a method based on differentiable rendering technology to generate large-scale, high-quality TDOM from UAV data without prior camera pose information. We progressively divide the scene into 4 blocks and use a 2DGS-based algorithm for parallel training on 4 GPUs. After fusing the point clouds, we perform batch rasterization to render orthophotos of the entire scene while simultaneously generating the corresponding depth maps. The results show that our method adapts well to different scenes, accurately represents building occlusion relationships, and reconstructs thin structures. Our TDOM with accuracy comparable to commercial software, combined with depth information, not only meets the demands of high-precision surveying and mapping but also provides users with more comprehensive and accurate spatial data support across multiple fields, thereby enhancing work efficiency and decision-making quality. However, our method still requires significantly longer training time per scene compared to commercial software. In future research, our aim is to improve the reconstruction efficiency of our method and further explore ways to save training resources.

\section*{Declarations}
\subsection{Funding}

The authors did not receive support from any organization for the submitted work.
All authors certify that they have no affiliations with or involvement in any organization or entity with any financial interest or non-financial interest in the subject matter or materials discussed in this manuscript.

\backmatter

\bibliography{sn-bibliography}

\end{document}